\documentclass[10pt,twocolumn,letterpaper]{article}

\usepackage{wacv}
\usepackage{times}
\usepackage{epsfig}
\usepackage{subcaption,graphicx}
\usepackage{amsmath}
\usepackage{amssymb}

\usepackage{import}
\usepackage{epstopdf}
\usepackage{caption}
\usepackage{subcaption}
\usepackage{tabu, tabularx}
\usepackage{verbatim}
\usepackage{nicefrac}
\usepackage{color}

\definecolor{darkblue}{rgb}{0.0, 0.0, 0.55}
\definecolor{input}{rgb}{0.63, 0.79, 0.95}
\definecolor{conv1}{rgb}{0.96, 0.87, 0.7} 
\definecolor{maxpool}{rgb}{0.92, 0.3, 0.26}
\definecolor{dense}{rgb}{0.82, 0.62, 0.91}
\definecolor{relu}{rgb}{0.95, 0.64, 0.38} 
\definecolor{pink}{rgb}{0.752, 0.086, 0.705}
\definecolor{blue}{rgb}{0.000, 0.455, 0.851}
\definecolor{green}{rgb}{0.239, 0.600, 0.439}
\definecolor{red}{rgb}{0.859, 0.078, 0.234}
\DeclareRobustCommand{\legendsquare}[1]{\textcolor{#1}{\rule{3ex}{1.5ex}}}

\wacvfinalcopy 

\def\wacvPaperID{1081} 

\ifwacvfinal\fi
\begin{document}

\setlength{\belowcaptionskip}{-\baselineskip}\addtolength{\belowcaptionskip}{1.1ex}

\title{Self-Attention Network for Skeleton-based Human Action Recognition}

\author{Sangwoo Cho, \hspace{0.3cm} Muhammad Hasan Maqbool, \hspace{0.3cm} Fei Liu, \hspace{0.3cm} Hassan Foroosh \\
Computer Science Department\\
University of Central Florida, Orlando, FL 32816, USA\\
{\tt\small \{swcho, hasanmaqbool\}@knights.ucf.edu, \{feiliu, foroosh\}@cs.ucf.edu}
}

\maketitle
\ifwacvfinal\thispagestyle{empty}\fi

\begin{abstract}
    Skeleton-based action recognition has recently attracted a lot of attention. Researchers are coming up with new approaches for extracting spatio-temporal relations and making considerable progress on large-scale skeleton-based datasets. Most of the architectures being proposed are based upon recurrent neural networks (RNNs), convolutional neural networks (CNNs) and graph-based CNNs. When it comes to skeleton-based action recognition, the importance of long term contextual information is central which is not captured by the current architectures. In order to come up with a better representation and capturing of long term spatio-temporal relationships, we propose three variants of Self-Attention Network (SAN), namely, SAN-V1, SAN-V2 and SAN-V3. Our SAN variants has the impressive capability of extracting high-level semantics by capturing long-range correlations. We have also integrated the Temporal Segment Network (TSN) with our SAN variants which resulted in improved overall performance.  
	Different configurations of Self-Attention Network (SAN) variants and Temporal Segment Network (TSN) are explored with extensive experiments. Our chosen configuration outperforms state-of-the-art Top-1 and Top-5 by 4.4$\%$ and 7.9$\%$ respectively on Kinetics and shows consistently better performance than state-of-the-art methods on NTU RGB+D.
\end{abstract}

\section{Introduction}
\begin{figure}[!th]
	\centering
	\includegraphics[width=\linewidth]{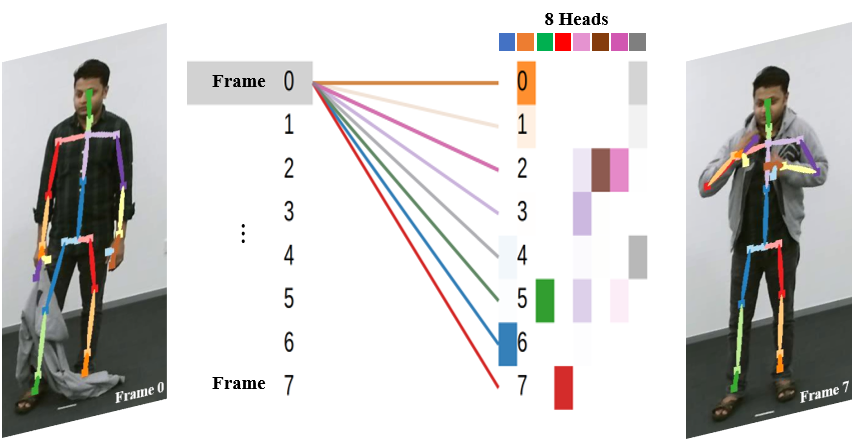}
	\caption{
		An example of self-attention response from the last self-attention layer.
		Eight frames are uniformly sampled from an action with the class `put on jacket' and illustrated as frame 0 to 7. Frame 0 has the strongest correlation with the last frame, frame 7, at the fourth head \legendsquare{red}, and attends heavily itself at the second head \legendsquare{relu}.
		Note that with the self-attention network each frame is associated with other frames so that local and global context information can be acquired.
	}
	\label{fig:teaser}
\end{figure}

Video-based action recognition has been an active research topic due to its important practical applications in many areas, such as video surveillance, behavior analysis, and video retrieval. Human action recognition can also be applicable to human-computer interaction or human-robot interaction to help machines understand human behaviors better \cite{intro1, intro2, cho2018temporal}. 
Unlike a single image that contains only spatial information, a video provides additional motion information as an important cue for recognition. 
Although a video provides more information, it is non-trivial to extract the information due to a number of difficulties such as viewpoint changes, camera motions, and scale variations, to name a few. 
There has been extensive research in RGB video-based action recognition and one of the mainstream methods is to employ both temporal optical flow and spatial appearance to obtain spatial and temporal information \cite{intro_2stream} . 
The RGB video datasets typically contain an extensive amount of data to process, hence require large models and resources to train them properly.
On the other hand, skeleton based action recognition comprises of only key joint locations of human bodies. 
With the advent of cost-effective depth cameras \cite{intro_kinects}, stereo cameras, and the advanced techniques for human pose estimation \cite{data:Openpose}, the cost to obtain key points has reduced and skeleton-based human action recognition has garnered increasing attraction \cite{intro_review1, STOA:H-RNN, STOA:stgcn}.
Although the key joint locations dont include appearance information, humans are able to recognizing actions from the motion of a few human skeleton joints according to Johansson \cite{intro_Johansson1973}. 
In this paper, we focus on human action recognition based on 3D skeleton sequences.

To extract information from skeleton sequences, many works naturally apply recurrent neural networks (RNNs) to model temporal dynamics \cite{STOA:PA-LSTM, STOA:TG_ST-LSTM, STOA:VA-LSTM}. They also utilize CNNs to model spatio-temporal dynamics by treating the 3D skeleton data as 2D pseudo images with 3 channels \cite{STOA:HCN, intro_cnn}. Another method is to retrieve structure information of human body by constructing a graph with human joints as edges \cite{STOA:stgcn}, which also based on CNNs.  
Despite the significant improvements in performance, there exist a problem to be solved.
Both recurrent and convolutional operations are neighborhood-based local operations \cite{intro_non-local} either in space or time; hence local-range information is repeatedly extracted and propagated to capture long-range dependencies.
Many works have designed networks with hierarchical structure \cite{STOA:H-RNN, STOA:ENS_TS-LSTM, cho2018spatio} to obtain longer range and deeper semantic information but the problem still persists if there are back and forth semantic dependencies.

\begin{figure}[!t]
	\centering
	\includegraphics[width=\linewidth]{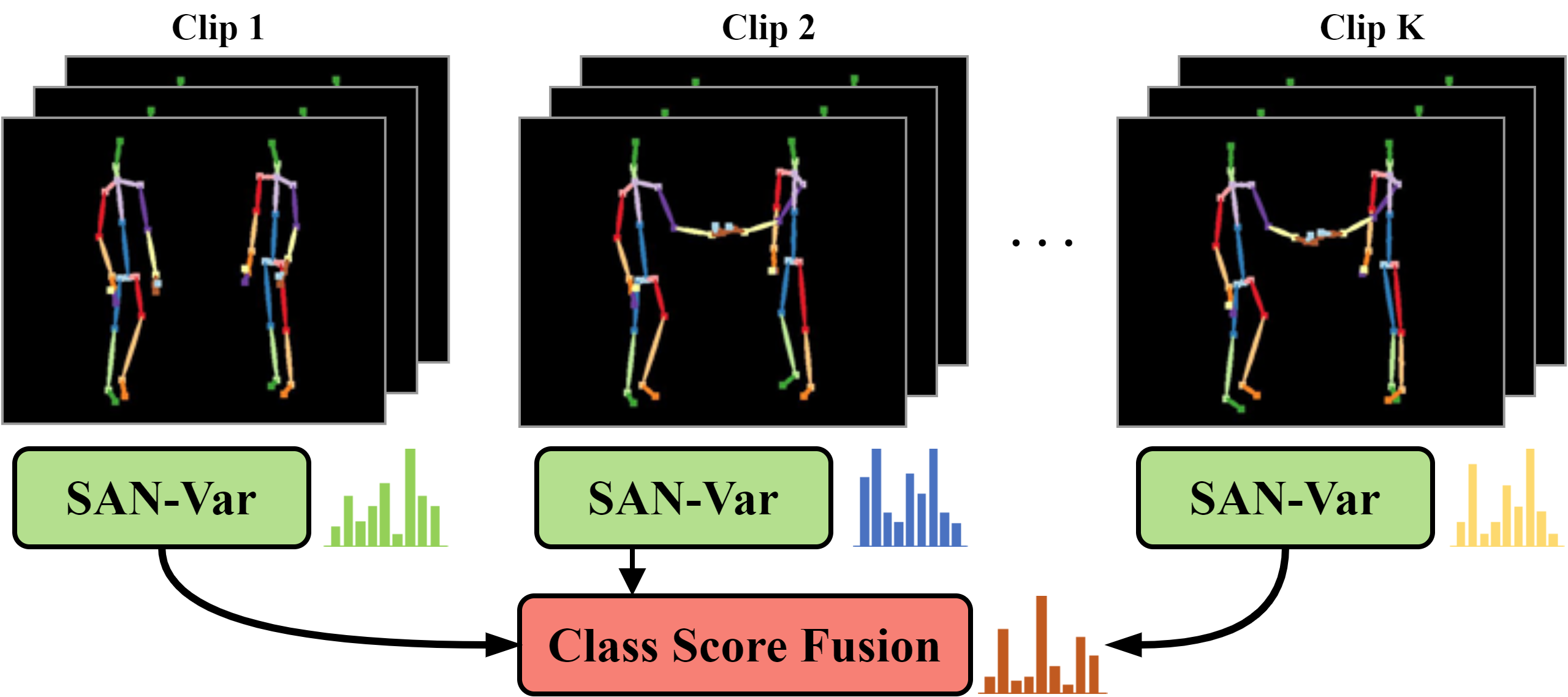}
	\caption{The overall pipeline of the proposed model. The network takes as inputs temporally segmented clips and extracts contextual information from each snippet by one of SAN variants described in section \ref{sec:SAN-base}. Predictions of each snippet are fused to compute the final prediction.
	}
	\label{fig:overall}
\end{figure}

In this paper, we propose a novel model with Self-Attention Network (SAN) to overcome the above limitation and retrieve better semantic information (Fig. \ref{fig:teaser}). 
Fig. \ref{fig:overall} shows the overall pipeline of our model.
The framework is motivated by temporal segment network \cite{intro_TSN} that extracts short-term information from each video sequence.
Our model extracts semantic information from each video sequence by SAN variants.
SAN-Variants take a sequence of features from encoded signals and computes the response at each position as a weighted sum of features at all positions. This operation enables SAN-Variants to correlate features in distance or even in opposite direction. 
The predicted outputs based on each clip are merged with consensus operations to capture deeper semantic understanding.
Therefore, our model can effectively solve the problem of acquiring long-term semantic information. 
Experimental results show that the learned SAN variants outperforms state of the art methods on challenging large scale datasets. 
We also visualize the attention correlations trying to understand how the network works and provide some insights.
The main contributions of the paper are summarized as follows:
\begin{enumerate}
    \itemsep-0.49em 
	\item We propose Self Attention Network (SAN) variants SAN-V1, SAN-V2 and SAN-V3 for effectively capturing deep semantic correlations from action sequences involving human skeleton. 
	\item We have integrated the Temporal Segment Network (TSN) with our SAN variants. We observed improved performance because of this integration of TSN and SAN variants.
	\item We visualize self-attention probabilities to show how each frame is correlated with other frames. 
	\item Our  proposed  method  achieves  state-of-the-art  results on two large scale datasets:  NTU RGB+D and Kinetics-skeleton
\end{enumerate}


\section{Related Work}
Handcrafted features are used to represent the skeleton motion information in early works. \cite{related_covariance} computes covariance matrix for joint positions over time. \cite{related_geometry} extracts 3D geometric relationships of body parts in Lie group based on rotations and translations of joints. With further progress in deep learning, researchers started using Recurrent Neural Networks to extract temporal dynamics between joints as RNNs use sequential processing. \cite{STOA:H-RNN} proposes a hierarchical RNN that splits the human body into five parts with each part fed into different subnetworks and fuses them hierarchically. \cite{STOA:PA-LSTM} splits a cell in an LSTM into part based cells and human body parts are applied to each cell to learn a representation of each part over time. \cite{related_co-occur} proposes a spatio-temporal LSTM network that learns the co-occurrence features of skeleton joints with a group sparse regularization. \cite{STOA:TG_ST-LSTM} introduces trust gate to reduce the influence of noisy joints and employs a spatio-temporal LSTM network to explore the spatila and temporal relationships. \cite{STOA:STA-LSTM} introduces attention mechanism in the LSTM network to focus on more important joints at each time instances.
In recent works, CNN based approaches \cite{related_conv1, related_conv2, related_conv3, related_conv4} are adopted to learn skeleton features and achieves significant performance. They attempt to convert a skeleton sequence into pseudo images and utilize CNNs to learn. \cite{related_conv2} maps a skeleton sequence to a tensor with frames, joints, and xyz coordinates treating it as image and leverages CNNs to train. \cite{related_conv1} proposes a method to use relative positions between the joints and the reference joints based on CNNs. \cite{related_conv4} maps trajectories of joints to orthogonal planes by using the 2D projection.
CNNs are also employed in our method to obtain more informative features from the raw skeleton joints. However, while the aforementioned RNNs and CNNs lack the ability to extract long-term correlation between features, our proposed method fills the gap to obtain high-level semantic information with long-range connections of features.

A self-attention network learns to generate hidden state representations for a sequence of input symbols using a multi-layer architecture~\cite{related_transformer}. 
The hidden states of the upper layer are built from the hidden states of the lower layer using a self-attention mechanism.
It learns to aggregate information from lower layer hidden states according to their similarities to the $t$-th hidden state.
The learned representations are highly effective because they capture deep contextualized information of the input sequence.
The self-attentive network with multi-head attention has demonstrated success on a number of tasks including machine translation~\cite{related_transformer,Tang:2018}, language modeling and natural language inference~\cite{related_BERT}, semantic role labeling~\cite{Strubell:2018}, often surpassing recurrent neural networks by a substantial margin. 
Particularly,~\cite{related_transformer} describes the Transformer model that makes the self-attention mechanism an integral part of the architecture for improved sequence modeling. 
\cite{related_BERT} learns deep contextualized word representations that have led to state-of-the-art performance on question answering and natural language inference without task-specific architecture modifications. 
Despite the success, self-attentive networks have been less investigated for the task of skeleton-based action recognition. In this paper, we introduce a novel self-attentive architecture to fill this gap.

Temporal information can be extracted from a sequence data or a video. Many research endeavors have introduced methods for modeling the temporal structure for action recognition \cite{related_ar1, related_ar2, related_ar3}. \cite{related_ar1} proposes to employ latent variables to decompose complex actions in time and \cite{related_ar2} introduces a latent hierarchical model that extends the temporal decomposition of complex actions. \cite{related_ar3} utilizes a rank SVM to model the temporal evolution of BoVW representations. \cite{intro_TSN} introduces a method to model a long-range temporal structure by simply splitting a video into snippets and fusing CNN outputs from each part. We adopt this method since it effectively extracts long-range temporal information and also is applicable to any network with end-to-end training.

\section{Self-Attention Network}
\begin{figure*}[!t]
	\centering
	\subcaptionbox{SAN-Block\label{fig:san_block}}{\includegraphics[width=0.85in]{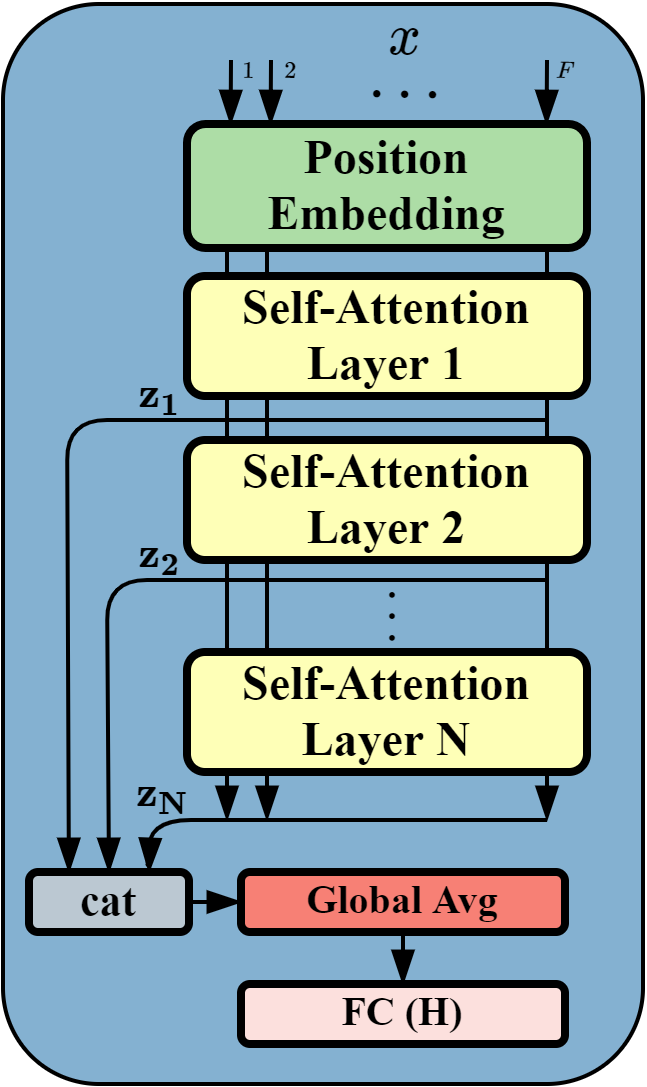}}\hspace{0.4em}%
	\subcaptionbox{SAN-V1\label{fig:base1}}{\includegraphics[width=0.8in]{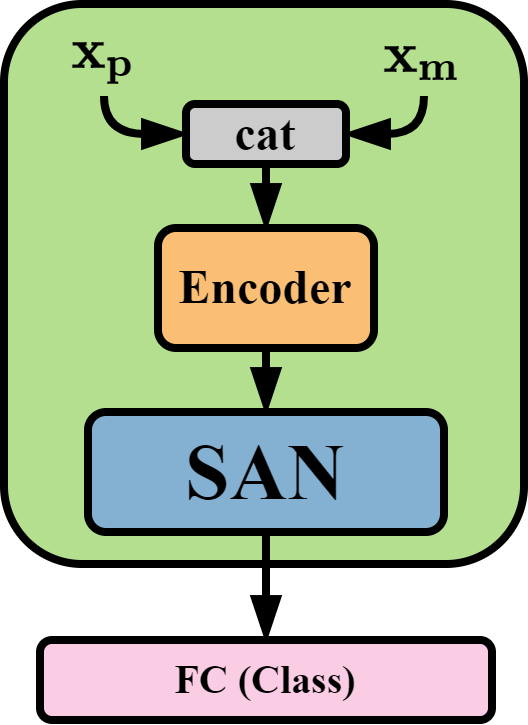}\vspace{0.83cm}}\hspace{0.6em}%
	\subcaptionbox{SAN-V2\label{fig:base2}}{\includegraphics[width=2.5in]{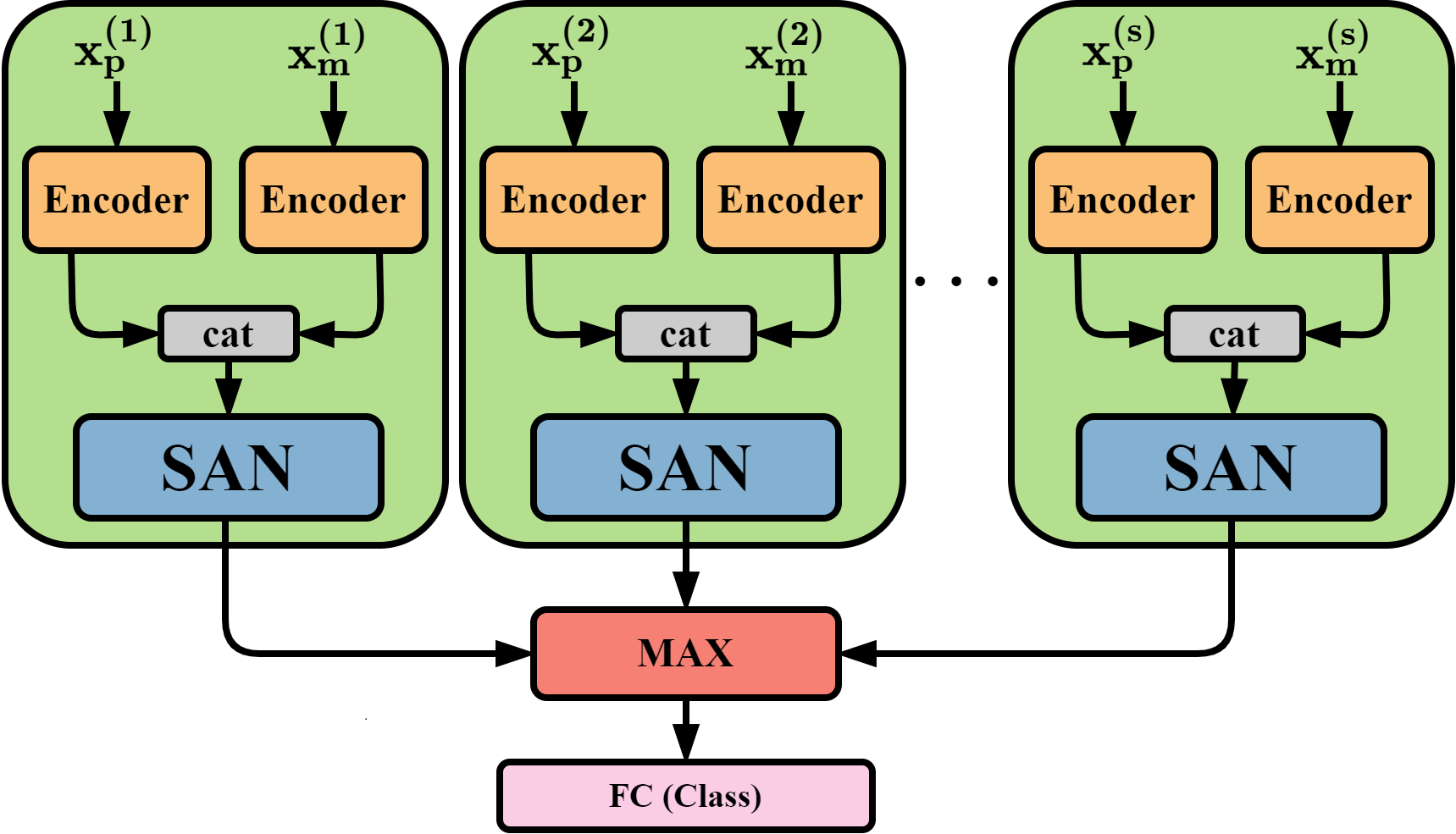}}\hspace{0.6em}%
	\subcaptionbox{SAN-V3\label{fig:base3}}{\includegraphics[width=2.5in]{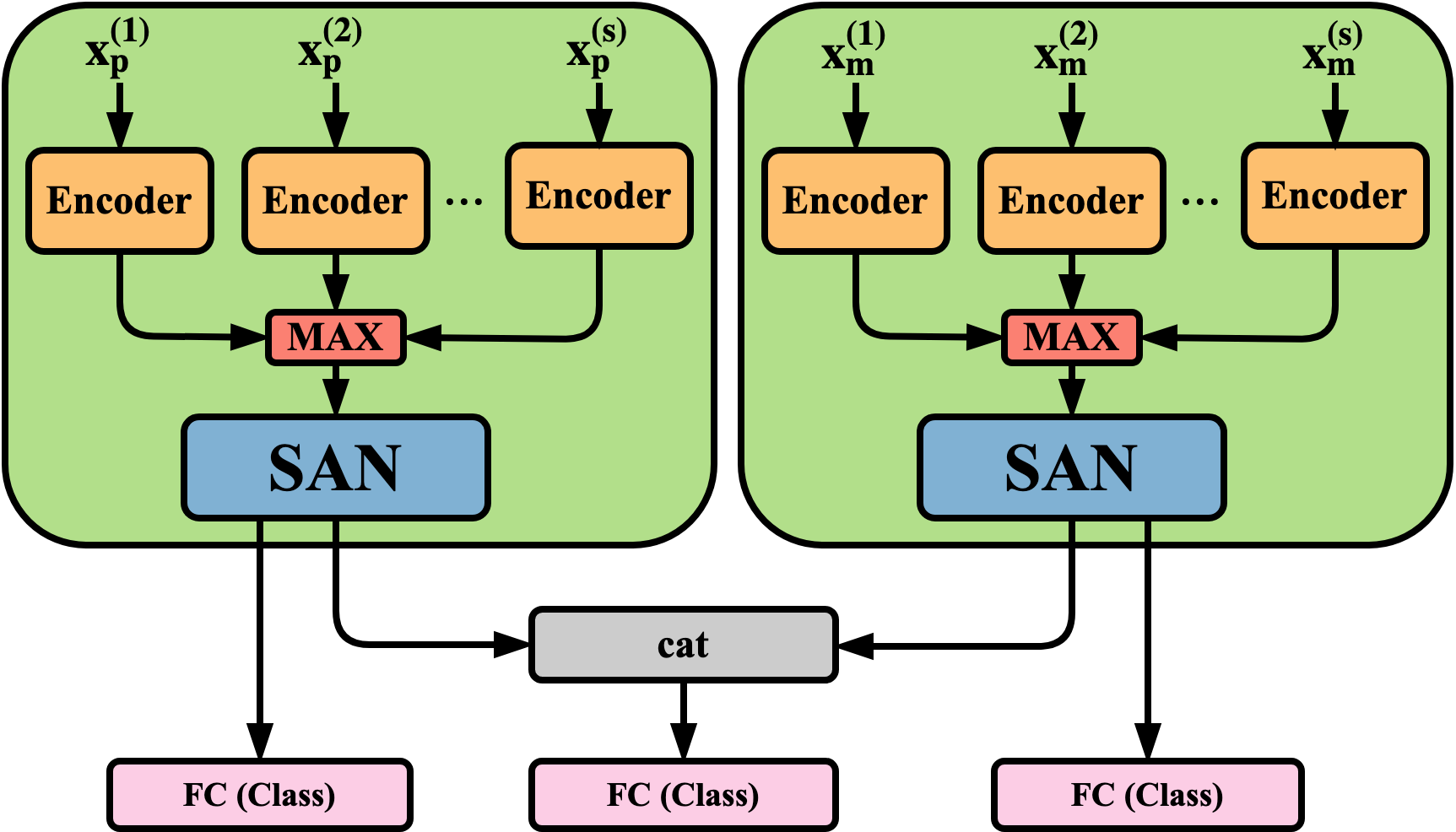}}
	\caption{
		Different designs of Self-Attention Network architecture. (a) self-attention network block (SAN) computing pairwise correlated attentions; (b) baseline model with early fused input features; (c) model that learns movements of each person in a scene; (d) model that learn different modalities for available people in a scene.
	}
	\label{fig:SAN_design}
\end{figure*}

In this section, we briefly review the Self-attention network.
Self-attention network \cite{related_transformer} is a powerful method to compute correlation between arbitrary positions of a sequence input. 
An attention function consists of a query $\textsf{A}_Q$, keys $\textsf{A}_K$, and values $\textsf{A}_V$ where query and keys have same vector dimension $\textsf{d}_k$, and values and outputs have same size of dimension $\textsf{d}_v$. The output is computed as a weighted sum of the values, and the weight assigned to each value is computed by scaled dot-product of query and keys. 
The vectors of query $\textsf{A}_Q$, keys $\textsf{A}_K$ and values $\textsf{A}_V$ are packed in a matrix generating $\textsf{Q}$, $\textsf{K}$, and $\textsf{V}$ matrices. Then the attention function is defined as 
\begin{equation}
Attention \left( \textsf{Q}, \textsf{K}, \textsf{V} \right) = softmax \left( \frac{\textsf{Q} \textsf{K}^T}{ \sqrt{\textsf{d}_k} } \right) \textsf{V},
\label{equ:attn}
\end{equation}
where $\frac{1}{\sqrt{\textsf{d}_k}}$ is a scaling factor.
The equation computes scaled dot-product attention and the network computes the attention multiple times in parallel (multi-head) to extract different correlation information. The multi-head attention outputs are concatenated and transformed to the same vector dimension the input sequence. A residual connection is adopted to take the input and output of the multi-head self-attention layer and a layer normalization is applied to the summed output.
A fully-connected feed-forward network with a residual connection is applied to the normalized self-attention output. The entire network is illustrated as a self-attention layer in Fig. \ref{fig:san_block} and multiple layers are repeated to extract better representation.

\section{Approach}
\label{sec:approach}
In this paper, we propose an effective model for skeleton-based action recognition, which is based on Self-Attention Network. The overall framework of the model is shown in Fig. \ref{fig:overall}. Primarily we have position and motion of joints. We can use raw position of the joints for figuring out the motion/velocity of the joints. Our SAN variants operate on encoded representations of position and motion sequences. We will be using simple non-linear projection (FCNN) and CNN based encoders for encoding the raw position and velocity sequences. First we will explain the data transformation from raw sequences of position and motion of the joints to encoded features. Once features are encoded, we will make use of three different SAN based architectures for effectively capturing the contextual information from the encoded features. 
\subsection{Raw Position and Motion Data }
The raw skeleton position $\mathbf{x}_\mathbf{p} \in \mathbb{R}^{F \times J' \times C}$ in a video clip is defined with the number of frames $F$, the number of joints per person $J$, and the coordinates of each joint $C$. There may be $S$ skeletons in a frame so the total number of joints is $J' = S \times J$. The position data can be depicted for each person as $\mathbf{x}_\mathbf{p}^{(s)}$, where $s \in \{1, 2, \cdots, S\}$.

The motion or velocity data, $\mathbf{x}_\mathbf{m} \in \mathbb{R}^{F \times J' \times C}$, can be explicitly retrieved by taking differences of each joint $J_{j}^t \in \mathbb{R}^{C}$, where $j \in \{1, \cdots, J\}$ and $t \in \{1, \cdots, F\}$, between consecutive frames:
\begin{equation}
\label{xm}
\mathbf{x}^t_\mathbf{m} = \left\{ J_1^{t+1}-J_1^t, J_2^{t+1}-J_2^t, \cdots, J_{\textsf{J}}^{t+1}-J_{\textsf{J}}^{t}  \right\}
\end{equation}
Similarly, the motion data for each person is represented as $\mathbf{x}^{(s)}_\mathbf{m}$.

\subsection{Encoder}
Our SAN variant models (Fig. \ref{fig:SAN_design}) operate upon the encoded position $\mathbf{x}_{\mathbf{(p,enc)}}$ and motion features $\mathbf{x}_{\mathbf{(m,enc)}}$. In this section, we describe two methods to encode the raw position $\mathbf{x}_\mathbf{p}$ and motion data $\mathbf{x}_\mathbf{m}$.

\subsubsection{Non-Linear Encoder}
\label{non-linear-encoder}
A non-linear encoder simply uses a feed-forward neural network (FCNN) with a non-linear activation function for projecting the input vector to higher dimension. 
For example, when encoding for SAN-V1 (Fig. \ref{fig:base1}) we perform early fusion of $\mathbf{x_p}$ and $\mathbf{x_m}$ to get $\mathbf{x} \in \mathbb{R}^{F \times 2J' \times C}$ and then use our non-linear encoder to get $\mathbf{x}_{\mathbf{(ff)}} \in \mathbb{R}^{F \times 2J' \times C'}$.  On the other hand, encoding for SAN-V2 (Fig. \ref{fig:base2}) and SAN-V3 (Fig. \ref{fig:base3}) individual skeletons are incorporated. 
In this case non-linear encoding is used to extend the skeleton joint position and motion tensor 
to $\mathbf{x}^{(s)}_{\mathbf{(p,ff)}} \in \mathbb{R}^{F \times J \times C'}$, and 
$\mathbf{x}^{(s)}_{\mathbf{(m,ff)}} \in \mathbb{R}^{F \times J \times C'}$, respectively.  

\subsubsection{CNN Based Encoder}
\label{CNN_block}
\begin{figure}[!t]
    \centering
	\begin{subfigure}[b]{.9\textwidth}
		\includegraphics[width=.5\textwidth]{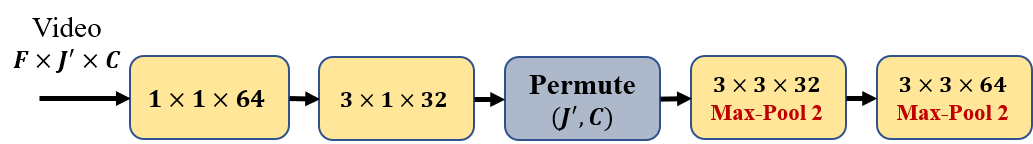}
		\label{fig:conv_block1}
	\end{subfigure}
	
	\centering
	\begin{subfigure}[b]{.9\textwidth}
	    \hspace{.4cm}
		\includegraphics[width=.5\textwidth]{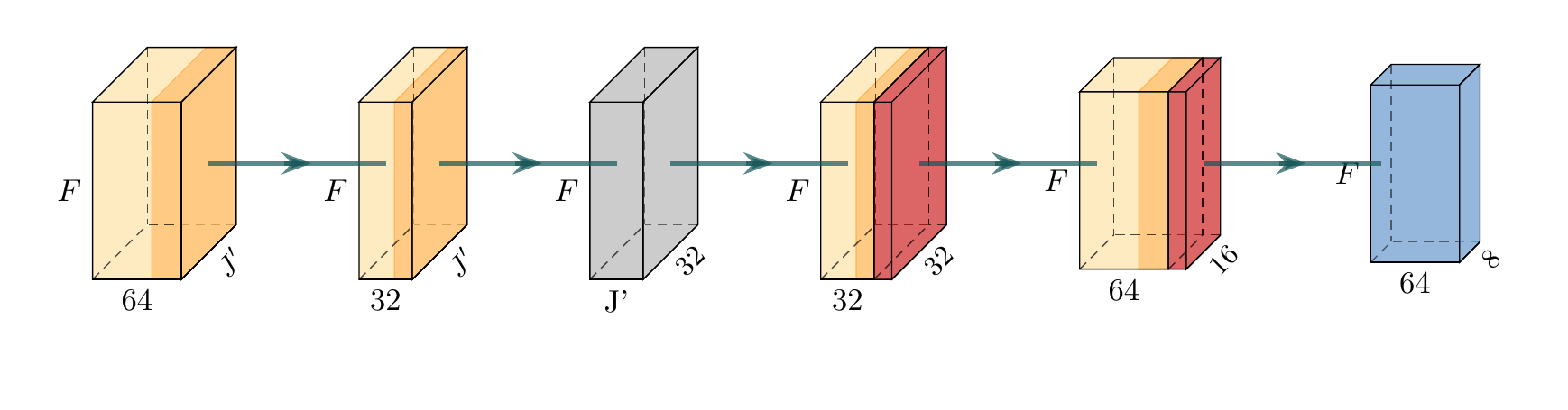}
		\label{fig:conv_block2}
	\end{subfigure}
	\caption{
		An input sequence of skeleton joints over frames, ${F \times J' \times C}$, is fed to the convolutional blocks and output tensor size of ${F \times 8 \times 64}$ is generated, which is denoted by \legendsquare{input}. Each color denotes the following layers: \legendsquare{conv1} convolutional layer; \legendsquare{relu} ReLU activation; and \legendsquare{maxpool} max-pooling layer.}
	\label{fig:convblock}
\vspace{-0.14in}
\end{figure}

A CNN based encoder is employed for encoding low level features from raw joint position and motion data $\mathbf{x}_\mathbf{p}$, $\mathbf{x}_\mathbf{m}$, or $\mathbf{x}^{(s)}_\mathbf{p}$, and $\mathbf{x}^{(s)}_\mathbf{m}$. 
2D convolutions can serve the purpose of extracting features from 3D tensors of raw skeleton data. Our encoder block consist of 4 convolutional layers as evident from Fig. \ref{fig:convblock}. 
We will explain the general encoding scheme by keeping in view the encoding requirements for SAN-V1 architecture. As we mentioned earlier in \ref{non-linear-encoder}, for SAN-V1, $\mathbf{x} \in \mathbb{R}^{F \times 2J' \times C}$ which is the output of early fusion of $\mathbf{x}_\mathbf{p}$ and $\mathbf{x}_\mathbf{m}$.
 First layer uses $1 \times 1 \times 64$ filters with stride $1$. Output of the first layer are the extended coordinates in the form of ${F \times J' \times 64}$ tensor. Layer two operates 
 with $3 \times 1 \times 32$ filters and stride $1$, and outputs a tensor of shape $F \times J' \times 32$. 
 Note that convolution window size for layer two is $3 \times 1$ because we are interested in extracting local contextual information over frames. Now, we transpose joints and cooridinates making the tensor of shape $F \times 32 \times J'$ in order to extract features from correlations of all joints over local frames.
 Third layer uses $3 \times 3 \times 32$ filters with stride 1 and max pooling with $1 \times 2$ pooling window is also applied. Output of third layer is a tensor with shape $F \times 16 \times 32$. Final convolution layer applies $3 \times 3 \times 64$ filters with stride 1. Similar to third layer, max pooling with a pooling window of $1 \times 2$ is also applied producing a $F \times 8 \times 64$ tensor. Last two CNN layers encode correlated local features from all joints of human body. For SAN-V2 (Fig. \ref{fig:base2}) and SAN-V3 (Fig. \ref{fig:base3}) we encode $\mathbf{x}^{(s)}_\mathbf{p}$ and $\mathbf{x}^{(s)}_\mathbf{m}$ for individual skeletons in the frames. Note that $F$ remains the same so feature representations for each frame are acquired with encoders.

\subsection{SAN Variant Architecture}
\label{sec:SAN-base}
We investigate three SAN based network architectures as shown in Fig.~\ref{fig:SAN_design} for skeleton based action recognition. These architectures employ the same SAN architecture as shown in Fig.~\ref{fig:san_block} but operate upon varying combinations of encoded features, $\mathbf{x}_{\mathbf{(enc)}}$, $\mathbf{x}_{\mathbf{(p,enc)}}$, and $\mathbf{x}_{\mathbf{(m,enc)}}$. 
We first discuss the SAN block used in the network in detail.

\subsubsection{Self-Attention Network}
SAN block operates on 
encoded representations of position and motion information. The input  to SAN block is $x \in \mathbb{R}^{F \times H}$, where $H$ is a feature representation per frame. The dimension of $H$ relys on the different encoders and model variants, and $H=512=8\times 64$ with the CNN encoder for SAN-V1.   
The first layer of the SAN block is a position embedding generating $p \in \mathbb{R}^{F \times H}$. Position embedding layer is used for providing a sense of order to the feature vectors. The ordering prior knowledge is helpful for each feature vector at each time to capture overall contextual cues from the input sequence.
The output of the position embedding layer $y$ is an element-wise addition of the input sequence $x$ and the position embedding $p$. 

Output of position embedding layer $y$ is fed to the first self-attention layer $\mathbf{z}_1$. Each SAN layer consumes the output of the previous SAN layer. Each self-attention layer computes pairwise attention probabilities and $K, Q $ and $V$ parameters described in Eq. \ref{equ:attn} are learned. 
Each self-attention layer outputs $\mathbf{z}_i, i \in \{ 1,2, \cdots, N \}$ where $N$ is the number of self-attention layers. We concatenate the outputs from each SAN layer in order to gather all the attention probabilities as shown below
\begin{align}
	c &= concat ( [ \mathbf{z}_1, \mathbf{z}_2, \cdots, \mathbf{z}_N  ] ) \\
	o &= \text{ReLU}(f_{lin} ( f_{avg} ( c ) ) )
\end{align}
where $concat$ layer concatenates $\mathbf{z}_i \in \mathbb{R}^{F \times H}$ along the vector axis creating a concatenated sequence $c \in  \mathbb{R}^{F \times HN}$. Then, a global average layer $f_{avg}$ is applied to $c$ along the frame axis to obtain video-level features and a resulting dimension of the feature is $\mathbb{R}^{HN}$. Finally, a fully connected layer $f_{lin}$ with a non-linear activation, ReLU, projects the feature vector to the same input dimension $H$.

\subsubsection{SAN-V1}
SAN-V1 (Fig. ~\ref{fig:base1}) is a baseline network to understand how well the SAN block works for this task. It takes a concatenated input of position ($\mathbf{x}_\mathbf{p}$) and motion ($\mathbf{x}_\mathbf{m}$) data generating an input sequence $\mathbf{x} \in \mathbb{R}^{F \times 2J'C}$. The concatenation is to achieve feature-level early fusion. $\mathbf{x}$ requires encoding which is achieved using CNN encoder and non-linear encoder. The shape of the input sequence to the encoders is $\mathbb{R}^{F \times H}$ where $H = 2 \times J' \times C$. 
SAN block extracts latent local and global context information out of the input encoded sequences $\mathbf{x_{conv}}$ and $\mathbf{x_{ff}}$.
Note that $J$ is the number of joints for one person, hence $J'$ represent the joints belonging to all the poeple in the frame. Zero paddings are applied in case that the number of valid people in a frame is less than a pre-defined maximum number of people.
The output of the SAN block is fed to a classification layer which consists of a ReLU activation layer, a dropout layer, and a linear layer with softmax activation to predict probabilities for each class. The network is trained with cross-entropy loss.

\subsubsection{SAN-V2}
SAN-V2 (Fig. \ref{fig:base2}) is designed to extract contextual features with the SAN blocks for each subject (skeleton) in a scene. This network computes actions for each skeleton and takes the strongest signal from all available people in a video.
Similar to SAN-V1, the encoded position and motion skeleton data for each person is concatenated respectively and the concatenated input sequences are fed to the corresponding SAN blocks.
The input dimension for each SAN block is $\mathbb{R}^{F \times 2JC'}$ and $\mathbb{R}^{F \times 2\times 512}$ with the non-linear and CNN encoder, respectively.
SAN blocks share weights to learn a variety of movements from different people. SAN outputs can be merged with different operations such as element-wise max, mean or concatenation. According to our preliminary experiments, element-wise max works the best as it captures the strongest action signal among people who may not be available. The final classification layer is identical to the one in SAN-V1.
Note that SAN-V2 leverages late fusion strategy and is scalable to arbitrary number of people.

\subsubsection{SAN-V3}
Lastly, SAN-V3 (Fig. \ref{fig:base3}) is designed to deal with different data modalities: position and velocity (or motion). 
The most prominent signals from all people are chosen by an element-wise max operation for each modality. The input dimension for the SAN block is $\mathbb{R}^{F \times JC'}$ and $\mathbb{R}^{F \times 512}$ for the non-linear and CNN encoder, respectively.
The output of each SAN block is fed to separate classifiers and the concatenated signal from the SAN blocks is consumed by another classifier.
This network is also scalable to any number of people in a scene. 
The training losses of the model are calculated by adding all cross entropy losses from each classifier.

\subsection{Temporal Segment Self-Attention Network (TS-SAN)}
The self-attention network can associate features in distance making it possible to capture long range information. However, as the feature representations for same action can vary with many constraints (viewpoint change, different speed of action by different subjects, etc), the proposed network may not learn well. 
Thus, we leverage the temporal segment network \cite{intro_TSN} to train the network more effectively. As shown in Fig. \ref{fig:overall}, a video is divided into $K$ clips and one of the SAN variants in Fig. \ref{fig:SAN_design} is employed to learn temporal dynamics on each clip. Note that all layers share weights for different clips.
Formally, given $K$ segments ${S_1, S_2, \cdots, S_K}$ of a video, the proposed network models a sequence of clips as follows:
\begin{multline}
	TS-SAN(S_1, S_2, \cdots, S_K) =  \mathcal{C} ( \mathcal{F}(S_1;\mathbf{W}), \mathcal{F}(S_2;\mathbf{W}),\\
	\cdots, \mathcal{F}(S_K;\mathbf{W})).
\end{multline}
where $\mathcal{F}$ denotes one of SAN-Variant models and $\mathbf{W}$ is its parameters. 
The predictions of each SAN model from each snippet are aggregated based on different function $\mathcal{C}$: element-wise max, and average. 

\section{Experiments}

\begin{table}[!t]
	\begin{center}
	\begin{tabular}{|c|c|c|}
		\hline
		Methods & CS & CV\\
		\hline
		H-RNN \cite{STOA:H-RNN} (2015) & 59.1 & 64.0\\
		PA-LSTM \cite{STOA:PA-LSTM} (2016) & 62.9 & 70.3\\
		TG ST-LSTM \cite{STOA:TG_ST-LSTM} (2016) & 69.2 & 77.7\\
		Two-stream RNN \cite{STOA:Two-stream_RNN} (2017) & 71.3 & 79.5\\
		STA-LSTM \cite{STOA:STA-LSTM} (2017) & 73.4 & 81.2\\
		Ensemble TS-LSTM \cite{STOA:ENS_TS-LSTM} (2017) & 74.6 & 81.3\\
		VA-LSTM \cite{STOA:VA-LSTM} (2017) & 79.4 & 87.6\\
		ST-GCN \cite{STOA:stgcn} (2018) & 81.5 & 88.3\\
		DPRL \cite{STOA:DPRL} (2018)     & 83.5 & 89.8\\
		HCN  \cite{STOA:HCN} (2018)     & 86.5 & 91.9\\
		SR-TSL  \cite{STOA:SR-TSL} (2018)     & 84.8 & 92.4\\
		\hline
		TS-SAN (Ours)  & \textbf{87.2}  & \textbf{92.7}  \\
		\hline
	\end{tabular}
	\end{center}
	\vspace{-0.15in}
	\caption{Results of our method in comparison with state-of-the-art methods on NTU RGB+D with Cross-Subject(CS) and Cross-View(CV) benchmarks.}
	\label{tab:ntud}
\end{table}

We perform extensive experiments to evaluate the effectiveness of our proposed Self-Attention frameworks on two large scale benchmark datasets: NTU RGB+D dataset \cite{data:NTUD}, and Kinetics-skeleton dataset \cite{data:Kinetics}. We analyze the performance of our variant models and visualize self-attention probabilities to understand its mechanism.
\subsection{Datasets}
\subsubsection{NTU RGB+D}
NTU RGB+D is the current largest action recognition dataset with joints annotations that are collected by Microsoft Kinect v2. It has 56,880 video samples and contains 60 action classes in total. These actions are performed by 40 distinct subjects. It is recorded with three cameras simultaneously in different horizontal views. The joints annotations consist of 3D locations of 25 major body joints. \cite{data:NTUD} defines two standard evaluation protocols for this dataset: Cross-Subject (CS) and Cross-View (CV). For Cross-Subject evaluation, the 40 subjects are split into training and testing groups. Each group consists of 20 subjects. The numbers of training and testing samples are 40,320 and 16,560, respectively. For Cross-View evaluation, all the samples of cameras 2 and 3 are used for training while the samples of camera 1 are used for testing. The numbers of training and testing samples are 37,920 and 18,960, respectively.
\subsubsection{Kinetics}
Kinetics \cite{data:Kinetics} contains about 266,000 video clips retrieved from YouTube and covers 400 classes. 
Since no skeleton annotation is provided, the skeleton is estimated by an OpenPose toolbox \cite{data:Openpose} from the resized videos of 340$\times$256 resolution.
The toolbox estimates 2D coordinates $(x,y)$ of 18 human joints and confidence scores $c$ for each joint. Each joint is represented as $(x,y,c)$ and 2 people are selected at most for each frame based on the highest average joint confidence score. The total number of frames for all clips is fixed to 300 by repeating the sequence from the start.
We employ the released skeleton dataset to train our model and report the performance of top-1 and top-5 accuracies as introduced in \cite{STOA:stgcn}. The numbers of training and validation samples are around 246,000 and 20,000, respectively. 

\subsection{Implementation Details}
 We resize the sequence length to a fixed number of $\mathbf{F}$=32/64 (NTU/Kinetics) with bilinear interpolation along the frame dimension. 
We use $K$=3 of temporal segments and 32 frames are sampled from each clip. The numbers of self-attention layers and multi-heads used for NTU RGB+D and Kinetics datasets are 4, 8 and 8, 8, respectively. 

To alleviate the problem of overfitting, we append dropout with a probability of 0.5 before the last prediction layer and after the last convolution layer. For the self-attention network, a 0.2 ratio of dropout is utilized. We employ a data augmentation scheme by randomly cropping sequences with a ratio of uniform distribution between [0.5, 1] for training. We center crop sequence with a ratio of 0.9 when testing.
The learning rate is initialized with $1e^{-4}$ and reduced by half in case no improvement of accuracy is observed for 5 epochs. Adam optimizer \cite{Adam} is applied with weight decay of $5e^{-5}$. The model is trained for 200/100 (NTU/Kinetics) epochs with a batch size of 64.

\subsection{Comparison to State of the Art}

\begin{table}[t]
	\begin{center}
	\begin{tabular}{|c|c|c|}
		\hline
		Methods & Top-1 & Top-5\\
		\hline
		Feature Enc. \cite{STOA:FeatureEnc} (2015) & 14.9 & 25.8\\
		Deep LSTM \cite{STOA:PA-LSTM} (2016) & 16.4 & 35.3\\
		Temporal Conv \cite{STOA:Temporal_Conv} (2017) & 20.3 & 40.0\\
		ST-GCN \cite{STOA:stgcn} (2018) & 30.7 & 52.8\\
		\hline
		TS-SAN (Ours) &  \textbf{35.1}  & \textbf{55.7}  \\
		\hline
	\end{tabular}
	\end{center}
	\vspace{-0.15in}
	\caption{Results of our method in comparison with state-of-the-art methods on Kinetics.}
	\label{tab:kinetics}
\end{table}

We compare the performance of the proposed method to the state-of-the-art methods on NTU RGB+D and Kinetics datasets as shown in Table \ref{tab:ntud} and Table \ref{tab:kinetics}. The compared methods are based on CNN, RNN (or LSTM), and graph structure and our method consistently outperform state-of-the-art approaches. This demonstrates the effectiveness of our proposed model for the skeleton-based action recognition task.

As shown in Table \ref{tab:ntud}, our proposed model achieves the best performance with 87.2$\%$ with CS and 92.7$\%$ with CV. Our model and \cite{STOA:STA-LSTM} have common in a sense that attention mechanism is used. By comparing with STA-LSTM \cite{STOA:STA-LSTM}, our model performs 13.8$\%$ with CS and 11.5$\%$ with CV. Our model encodes the raw skeleton data with CNNs similar to HCN \cite{STOA:HCN} but outperforms by 0.7$\%$ with CS and 0.8$\%$ with CV. Comparing our model with SR-TSL \cite{STOA:SR-TSL} which is one of the best-performed methods, the performance gaps are 2.4$\%$ with CS and 0.3$\%$ with CV.

On the Kinetics dataset, we compare with four methods which are based on handcraft features, LSTM, temporal convolution, and graph-based convolution. As shown in Table \ref{tab:kinetics}, our method attains the best performance with a significant margin. The proposed method outperforms by 4.4$\%$ on top-1 and 2.9$\%$ on top-5 accuracies.
We observe that CNN based methods \cite{STOA:HCN, STOA:SR-TSL, STOA:stgcn, STOA:Temporal_Conv} are superior to LSTM based methods \cite{STOA:VA-LSTM, STOA:ENS_TS-LSTM, STOA:PA-LSTM} based on both Table \ref{tab:ntud} and Table \ref{tab:kinetics}, and our model outperforms the CNN based methods.

\subsection{Ablation Study}
We analyze the proposed network by comparing it with baseline models. 
We compare SAN variants with hyperparameter options for encoders, self-attention network, and temporal segment network.
Each experiment is evaluated on the NTU RGB+D dataset.

\begin{table}[!t]
	\begin{center}
	\begin{tabular}{|c|c|c|}
		\hline
		Methods & CS & CV\\
		\hline
		SAN-V1 + FF  & 75.4 & 79.8\\
		SAN-V1 + CNN  & 80.1 & 86.2\\
		SAN-V2 + FF  & 80.3 & 85.2 \\
		SAN-V2 + CNN  & 85.9 & 91.7\\
		SAN-V3 + FF  & 78.6 & 84.1\\
		SAN-V3 + CNN  & 85.5 & 91.4\\
		\hline
	\end{tabular}
	\end{center}
	\vspace{-0.15in}
	\caption{The comparison results of SAN variants shown in Fig. \ref{fig:SAN_design} with different encoder inputs on NTU dataset ($\%$).}
	\label{tab:SAN-base-ablation}
\end{table}

\begin{table}[!t]
    \begin{center}
	\begin{tabular}{|c|c|c|}
		\hline
		Methods & CS & CV\\
		\hline
		SAN-V2 (seq=96)  & 86.1 & 92.0\\
		SAN-V3 (seq=96)  & 85.9 & 91.7\\
		TS (seg=3) + SAN-V2 (seq=32)  & 87.2 & 92.7\\
		TS (seg=3) + SAN-V3 (seq=32)  & 86.8 & 92.4\\
		\hline
	\end{tabular}
	\end{center}
	\vspace{-0.15in}
	\caption{The comparison results of effectiveness of temporal segment on NTU dataset ($\%$).}
	\label{tab:TS-SAN-ablation}
\end{table}

\subsubsection{Effect of SAN Variants with Different Encoders}
Table \ref{tab:SAN-base-ablation} shows the results with different SAN variants and different inputs to them.
The SAN-V2 model performs the best and the SAN-V1 model the worst. The gap between the SAN-V2 model and the SAN-V3 model is minimal. We observe that the CNN encoder boosts the performance accuracy by up to 7.3$\%$ for SAN-V3. It shows that the CNN encoder effectively generates rich feature representations for the SAN models and plays a significant role in the network.
From the observation that SAN-V2 slightly outperforms SAN-V3, we conclude two facts: late fusion performs better than early fusion; and sharing weights of SAN blocks resulting in better trained models.

\subsubsection{Effect of Temporal Segment}
The self-attention network is suitable for connecting both short and long-range features and is capable of capturing higher-level context from all correlations. We compare the TS-SAN and SAN variants to see how they perform differently if two networks have the same sequence length. 
As shown in Table \ref{tab:TS-SAN-ablation}, TS-SAN outperforms. 
This proves that our design goal to make use of the temporal segment is correct. However, the SAN variants without the temporal segment network have an advantage of having less parameters with a small sacrifice of performance.
Although TS-SAN models outperform, we observe that the SAN variants perform well for long-range input sequences, $F$=96.

\begin{table}[!t]
	\begin{center}
	\begin{tabular}{|c|c|c|}
		\hline
		Methods & CS & CV\\
		\hline
		TS(Avg) + SAN-V2 & 87.2  & 92.7\\
		TS(Max) + SAN-V2 & 86.1  & 91.9\\
		TS(Avg) + SAN-V3 & 86.8  & 92.4\\
		TS(Max) + SAN-V3 & 85.9  & 91.1\\
		\hline
	\end{tabular}
	\end{center}
	\vspace{-0.15in}
	\caption{The comparison results of different aggregation methods for TS network on NTU dataset ($\%$).}
	\label{tab:TS-agg-ablation}
\end{table}
\begin{table}[!t]
	\begin{center}
	\begin{tabular}{|c|c|c|}
		\hline
		Methods & CS & CV\\
		\hline
		TS + SAN-V2 (L2H2) & 86.7 & 92.1\\
		TS + SAN-V2 (L4H4) & 86.9 & 92.5\\
		TS + SAN-V2 (L4H8) & 87.2 & 92.7\\
		TS + SAN-V2 (L8H8) & 87.0 & 92.4\\
		\hline
	\end{tabular}
	\end{center}
	\vspace{-0.15in}
	\caption{The comparison results of the number of attention layers and multi-heads on NTU dataset  ($\%$).}
	\label{tab:SAN-LH-ablation}
\end{table}

\subsubsection{Effect of Consensus Function}
We consider element-wise operations for the consensus function to compute the final prediction.
Two operations are valid: element-wise average, element-wise maximum.
Table \ref{tab:TS-agg-ablation} shows the performances of TS-SAN-V2 and TS-SAN-V3 with the above operations.
The element-wise average consensus function outperforms the element-wise max operation in both SAN variants.
The TS-SAN model with the element-wise max operation is outperformed by the SAN model without the temporal segment as shown in Table \ref{tab:TS-SAN-ablation}.
We conjecture that since the self-attention output signals are based on weighted average computation, it makes more sense to use the element-wise average aggregation function for the collected outputs from each snippet. By doing so, the video level self-attention can be computed properly leading to the best performance.

\subsubsection{Effect of Number of Layers and Mutli-heads in SAN Block}
We compare TS-SAN-V2 model with different number of layers and multi-heads. The results are shown in Table \ref{tab:SAN-LH-ablation}. By comparing the row 2 and 3, we observe that the number of heads affect the performance marginally. 
From the results of the row 3 and 4, we also observe that the network underperform if it contains too many paramerters. 
On the contrary, the network also underperforms when the number of parameters are not enough (row 1).
According to the results, we argue that the proposed model requires a proper number of layers and heads for a cetrain dataset to perform the best.


\subsection{Visualization of Self-Attention Layer Response}

\begin{figure}[!t]
	\centering
	\subcaptionbox{`Put on jacket' action with subject 1\label{fig:vis1}}{\includegraphics[height=2.7cm,width=\linewidth]{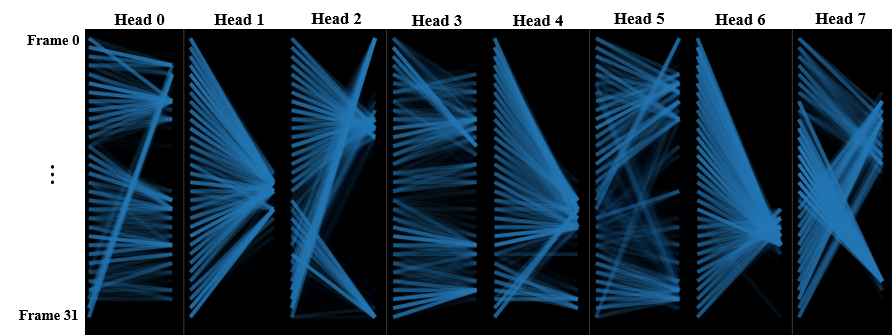}}\vspace{0.7em} 
	
	\subcaptionbox{`Put on jacket' action with subject 2\label{fig:vis2}}{\includegraphics[height=2.7cm,width=\linewidth]{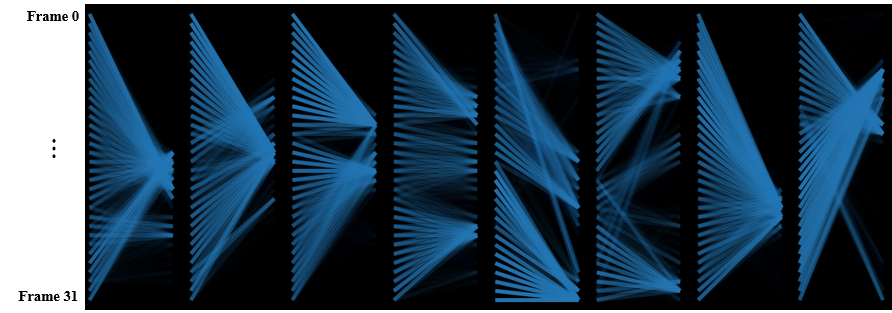}}\vspace{0.7em}%
	
	\subcaptionbox{`Reading' action with subject 1\label{fig:vis3}}{\includegraphics[height=2.7cm,width=\linewidth]{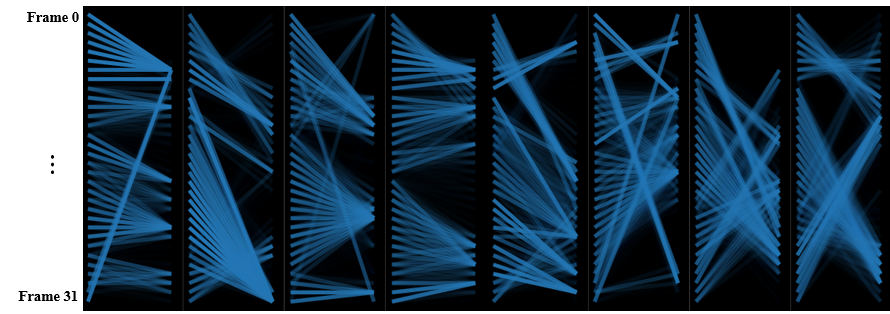}}
	\caption{
		Self-attention probabilities from the last self-attention layer for three test videos on NTU RGB+D are visualized. 
		The brighter color denotes the higher probability or the stronger connection. 
	}
	\label{fig:vis}
\end{figure}

The self-attention network determines where each frame correlates to other frames.
We visualize the self-attention response from the last self-attention layer with a visualization tool~\cite{vis} to understand how each frame is correlated for a certain action video.
As shown in Fig.~\ref{fig:vis}, the vertical axis shows the sampled 32 frames. Self-attention responses for eight multi-heads are displayed and each column shows the coarse shape of the attention pattern between two frames.

The model used for this visualization attains four layers and eight heads, and takes 32 sampled frames as the input sequence. No temporal segment network is used to train the network.
The self-attention probabilities are calculated by the equ. \ref{equ:attn} in the self-attention layer described in Fig.~\ref{fig:san_block}.
For example, from Fig. \ref{fig:vis1}, one of the strongest correlation in the third head can be found from a connection between frame 31 to frame 0 (a line across from bottom left to top right). 
From the above example, we can check the long range correlation is achieved and the proposed method captures a variety of correlations in both short and long distance.

We observe that the overall self-attention response patterns of the same action class (`put on jacket') resembles each other as shown in Fig. \ref{fig:vis1} and Fig. \ref{fig:vis2}. The repsonses of head 1 and head 6 from two videos especially shows similar pattern. Although two videos are taken by different subjects, duration, and views, we can see that the self-attention catches a certain latent similarity.
Comparing Fig. \ref{fig:vis1} and Fig. \ref{fig:vis2} with Fig. \ref{fig:vis3}, there is not much similar response pattern between them due to different action classes (`put on jacket' vs `reading').
We also learn that the proposed model is robust to subtle motion or speed of action changes from difference subjects or even views. 

\section{Conclusion}
In this paper, we propose three novel SAN variations in order to extract high-level context from short and long-range self-attentions. Our proposed architectures significantly outperform state-of-the-art methods.
CNN employed in our model is effective to extract feature representations for the input sequence of the self-attention network.
SAN can capture the temporal correlations regardless of distance, making it possible to obtain high-level context information from both short and long-range self-attentions.
We also propose an effective integration of SAN and TSN which results in observable performance boost.
We perform extensive experiments on two large scale datasets, NTU RGB+D and Kinetics-skeleton, and verify the effectiveness of our proposed models for the skeleton-based action recognition task. 
In the future, we will apply our model to video-based recognition tasks with key point annotations, such as facial expression recognition. We will also explore different methods to extract effective feature representations for the input sequence of SAN.

{\small
\bibliographystyle{ieee}
\bibliography{ref,fei_new}
}
\end{document}